\documentclass[letterpaper, 10 pt, conference]{ieeeconf}
\usepackage{graphicx} 

\usepackage{subcaption}
\usepackage{colonequals}
\usepackage{amsmath} 
\usepackage{float}
\usepackage{booktabs}
\usepackage{amsfonts}
\usepackage{amssymb}
\usepackage{xcolor}
\usepackage{siunitx}
\usepackage[font=small,labelfont=bf]{caption}
\DeclareMathOperator*{\argmax}{arg\,max}

\usepackage[ruled]{algorithm2e}
\makeatletter
\newcommand{\algorithmstyle}[1]{\renewcommand{\algocf@style}{#1}}
\newcommand{\removelatexerror}{\let\@latex@error\@gobble}
\makeatother

\IEEEoverridecommandlockouts                              

\title{\LARGE \bf
Evaluation of state representation methods in robot hand-eye \\coordination learning from demonstration
}

\author{Jun Jin$^{1}$, Masood Dehghan$^{1}$, Laura Petrich$^{1}$, Steven Weikai Lu$^{1}$ and Martin Jagersand$^{1}$
\thanks{$^{1}$Authors are with Department of Computing Science,
        University of Alberta, Edmonton AB., Canada, T6G 2E1
        {\tt\small jjin5@ualbertra.ca}}%
}

\begin{document}

\maketitle
\thispagestyle{empty}
\pagestyle{empty}

\begin{abstract}
We evaluate different state representation methods in robot hand-eye coordination learning on different aspects. Regarding state dimension reduction: we evaluates how these state representation methods capture relevant task information and how much compactness should a state representation be. Regarding controllability: experiments are designed to use different state representation methods in a traditional visual servoing controller and a REINFORCE controller. We analyze the challenges arisen from the representation itself other than from control algorithms. Regarding embodiment problem in LfD: we evaluate different method's capability in transferring learned representation from human to robot. Results are visualized for better understanding and comparison.
\end{abstract}
\section{Introduction}
In this paper, we discuss three problems when using state representation methods in robot hand-eye coordination learning by watching demonstrations, in which case, states are raw images: \textbf{(1) Regarding state dimension reduction:} How compact should a state representation be? Is it compact enough? Does this compact representation capture the whole task relevant information? How does each unit in the latent vector represent a task (disentangled representation~\cite{bengio2013deep})? \textbf{(2) Regarding controllability}: Can this state representation be used in a traditional hand-eye coordination controller (e.g., visual servoing~\cite{Chaumette2006}), as well as learning based control methods~\cite{sutton2018reinforcement}? What are the challenges arisen from state representation itself, other than from control algorithms? \textbf{(3) Regarding Embodiment Problem in LfD~\cite{Argall2009}:} Instead of meta-learning methods~\cite{yu2018one}, can this learned representation be directly used in a robot case (from human hand to robot end effector)?

\begin{figure}
	\begin{center}
		\includegraphics[width=0.48\textwidth]{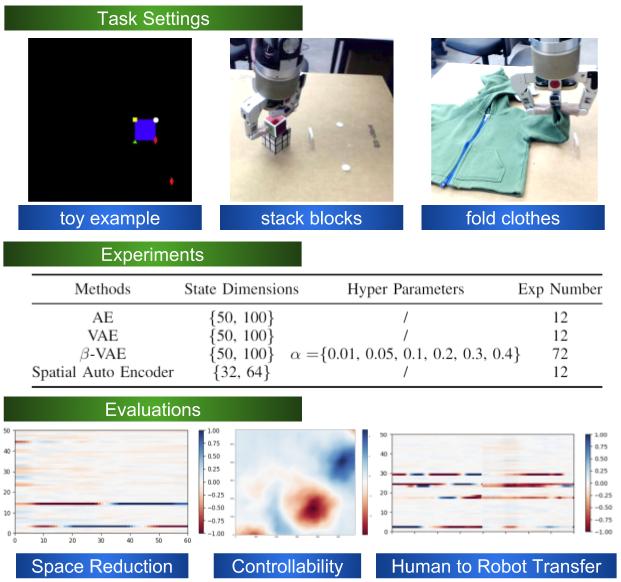} 
		\caption{We evaluate different state representation methods used in robot hand-eye coordination tasks. Our evaluation focus on (1) state space reduction: how is this compact representation relevant to the task?; (2) controllability: what are the challenges arisen from the representation itself in control; (3) embodiment problem: what if we directly transfer the learned representation from human to robot?}
		\label{fig:design_overview}
	\end{center}
\end{figure}
These problems are important in both traditional methods (e.g., visual servoing~\cite{Chaumette2006,Bateux2017}) and learning based methods (e.g., ~\cite{Levine2016,Finn2016}). For example, in image-based Visual Servoing (IBVS~\cite{Chaumette2006}), a task is defined by identifying target and current feature points~\cite{Gridseth2016}. The error between them is then stacked as a vector, which can be seen as a compact state representation in IBVS. The dimension of the original state, raw image, is greatly reduced by constructing this error vector. Control laws are derived by exponentially reducing this error~\cite{Chaumette2006}, thus guarantees task fulfillment. Hagger and Dodds et al. 2000~\cite{Hager2000} found that a task function can be generalized using geometric constraints~\cite{Hager2000,Gridseth2016} (e.g., point to point, point to line, etc.). Also, this constructed task function is a compact representation of the observed image. Factors in this compact representation are independent. Controlling over each factor will correspond to one aspect of fulfilling the task (e.g., reaching a target w.r.t. coordinate in x-axis)~\cite{Gridseth2016}. 

In learning based methods, for example, in the end to end learning paper~\cite{Levine2016}, a spatial autoencoder is used to represent a state image, which captures task relevant information (e.g, tool moving, target), thus defines a task. Furthermore, this compact representation brings another benefit in control (model-based RL~\cite{sutton2018reinforcement}). Since it's compact, it provides a way to approximate the unknown system dynamics and optimized all together in the final RL learning process~\cite{Levine2016,watter2015embed}. 

\begin{figure*}
	\centering
	\includegraphics[width=0.98\textwidth]{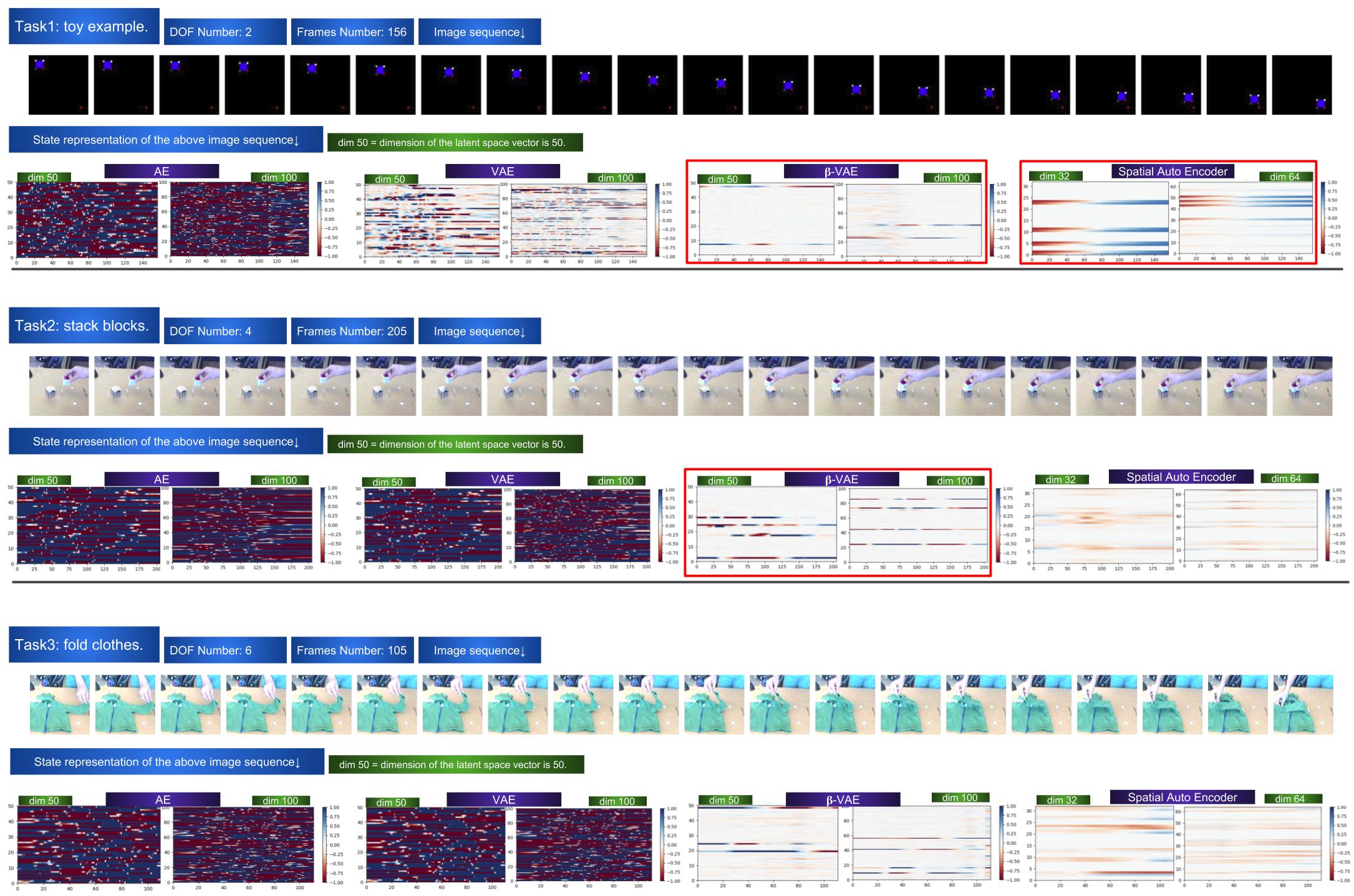}
	\caption{\textbf{task map}: Given a sequence of images recording task execution process, we visualize each method's latent space on different dimension settings. X-axis defines the time line, while y-axis defines value of each element in the latent vector. A task map visualize how latent vector changes along time. As shown in VAE, $\beta$-VAE and SAE methods, some factors are changing smoothly along time while some are fixed vibrating around zero. We further observes that the number of time-varying factors increases as task DOF number increases. And it remains the same even if dimension of latent vector increases. A larger latent space dimension (100) doesn't have much advantages over a smaller one (50).}
	\label{fig:net_design}
\end{figure*}

Various state representation methods~\cite{watter2015embed,kingma2013auto,higgins2017beta,Finn2016} are proposed in robot control, a review can be found in~\cite{lesort2018state}. Current evaluation of these methods are more on the general RL problem setting. In hand-eye coordination tasks, problems regarding how the representation is relevant to the task, what are the challenges arisen from the representation itself in a controller and how about transferring the representation from human to robot, are still interesting to explore.

In this paper, we experimentally evaluate 4 existing state representation methods in a hand-eye coordination task setting. Three tasks (Fig. 1) are designed over 4 existing state presentation methods: Autoencoder (AE~\cite{hinton2006reducing}), Variational Autoencoder(VAE~\cite{kingma2013auto}), $\beta$-VAE~\cite{higgins2017beta} and Spatial Autoencoder (SAE~\cite{finn2016deep}). Evaluations are done in three aspects: (1) state space reduction; (2) Controllability; (3) Representation transferring from human to robot. Our contributions are:
\begin{itemize}
    \item Experiments are conducted over 4 state representation methods in a robot hand-eye coordination task setting. Our analysis shows the effect of disentangling time-varying factors and time-fixed factors during a demonstration sequence.
    \item We evaluate controllability in both a visual servoing controller and REINFORCE~\cite{sutton2018reinforcement} controller in the toy example task. We visualize the mapping from task space to the latent space by sampling the whole task space. This visualization helps understanding challenges when using the learned state representation in control.
    \item At last, we evaluate each representation method's ability when transfer from human task setting to a robot task setting. This ability is crucial in task teaching, since learning by watching human demonstration provides the most intuitive way to specify a new task.
\end{itemize}
\begin{figure*}
	\centering
	\includegraphics[width=0.98\textwidth]{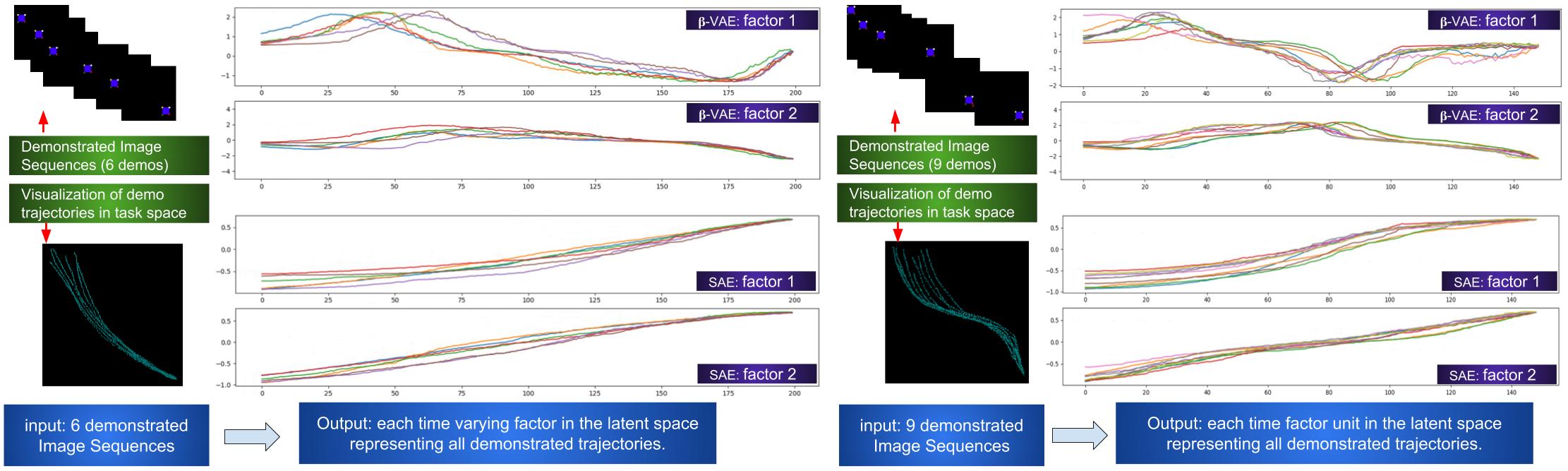}
	\caption{How time-varying factors represent trajectory patterns. Given several demonstrated trajectories with the same pattern, each trajectory is represented as one image sequence. These image sequences are then fed into $\beta$-VAE and spatial autoencoder, their corresponding time-varying factors (dim: 2) are computed and visualized along time axis. X-axis is the time line. Y-axis is time-varying factor's value. Two sets of trajectories are shown in this figure.}
	\label{fig:net_design}
\end{figure*}

\section{Methods}
\subsection{Problem formation}
A state representation problem in hand-eye coordination task setting can be format as: \textbf{Given} a set of human/robot demonstrations $\{\mathbf{\tau_{i}}\}$, where $\mathbf{\tau_{i}} = \{I_{0}, I_{1}, ..., I_{t}, ..., I_{T}\}$ and $I_{t}$ is a raw image captured during demonstration, \textbf{what} is a compact representation of $I_{t}$ that can best captures task relevant information for the purpose of fulfilling the task? The goal is to estimate a function $\phi$ that maps from $I_{t}$ to $s_{t}$:
\begin{equation}
\label{eq:11}
s_{t}=\phi(I_{t})
\end{equation} 
\subsection{State Representation Methods}
T. Lesort1 et al. 2018~\cite{lesort2018state} provide a comprehensive review of all existing state representation methods used in robotic control. Since most details are covered in literature, we will only give a brief review.
\subsubsection{Autoencoder}
An autoencoder(AE~\cite{hinton2006reducing}) learns minimizing the reconstruction error. The learned latent space has lower dimension than the input state (image) has. Other methods (e.g., DAE~\cite{vincent2010stacked}) are also proposed.

\begin{figure}
	\begin{center}
		\includegraphics[width=0.48\textwidth]{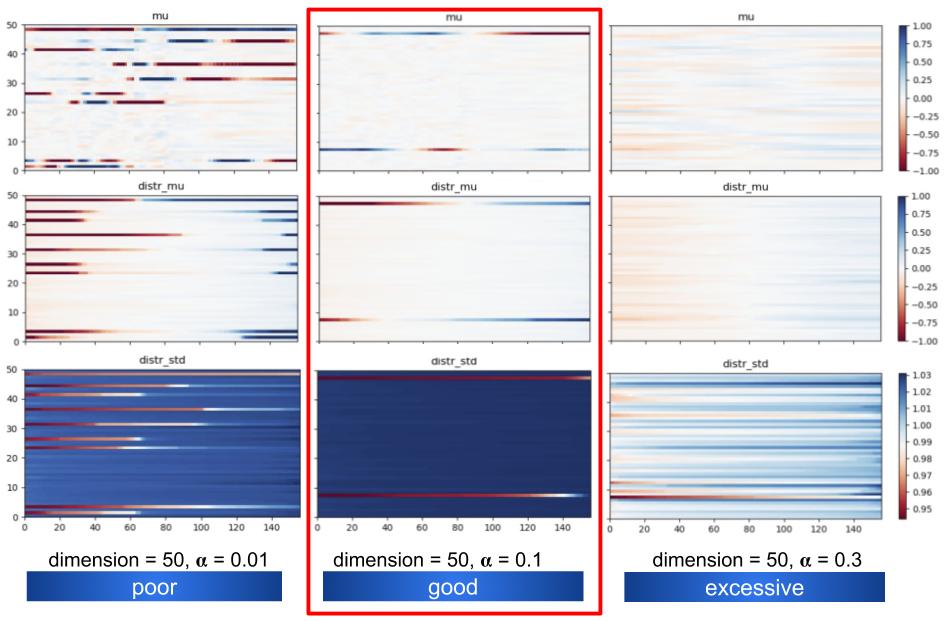} 
		\caption{Different hyper-parameter ($\alpha$) in $\beta$-VAE greatly affects the result. The first row shows how latent vector changes along the time line in a task execution process. The second row visualize the distribution of each unit in the latent vector. The last row shows how variance changed along the time line. Left column with a smaller $\alpha$ shows poor results in disentangling time-varying factors. Right column with a larger $\alpha$ shows excessive results since most factors are forced to vibrating around zero. And the middle column shows the best result.}
		\label{fig:design_overview}
	\end{center}
\end{figure}

\subsubsection{Variational Autoencoder}
A variational autoencoder (VAE~\cite{kingma2013auto}) learns based on mean-field variational inference~\cite{wainwright2008graphical} with a prior assumption that factors in the latent space are independent thus can be factorized. It enables the use of mean-field approximation to solve the problem. The objective is then converted to an Evidence Lower Bound (ELBO) which also minimize the reconstruction error, as well as the KL divergence between the learned latent space distribution and our prior assumption. This prior factorization assumption makes the learned factors independently distributed~\cite{higgins2017beta}, which enables a disentangled representation.

\subsubsection{$\beta$-VAE}
$\beta$-VAE~\cite{higgins2017beta} is proposed to further investigate the disentangling effect in the original VAE. The hyper parameter $\beta$ is often set to greater than 1 (e.g., 100 or more), which will push the optimization towards more to the mean-field prior and produces a better disentanglement result. ~\cite{higgins2017beta} reported the use of $\beta_{norm}$ as a better hyper parameter to deal with magnitude difference between input and latent vector sizes. Borrowing the same idea, we introduce $\alpha$ as our hyper parameter. From $\alpha$ we can compute $\beta$ using:
\begin{equation}
\label{eq:11}
\beta = \alpha \frac{dim\_input}{dim\_{z_{t}}},
\end{equation} 
where, $dim\_input$ is calculated by input image's $width*height*channels$ and $dim\_z_{t}$ is the dimension of latent vector.

\subsubsection{Spatial Autoencoder, SAE}
SAE~\cite{finn2016deep} is proposed to encode spatial information more than regular image density information. A spatial softmax is used tp output expected x and y positions of each softmax probability distribution. It plays a similar role of trackers when compared to feature based IBVS~\cite{Chaumette2006}.

\section{Experiments and Evaluations}
\subsection{Tasks}
We designed three tasks in our experiments (Fig. 1): \textbf{(1)} a toy example which provides a noise-free environment, enables sampling over the whole task space and has no embodiment problem; \textbf{(2)} a stack blocks task which has regular rigid geometric shapes; \textbf{(3)} a fold clothes task which has deformable shapes during execution. Each task contains several demonstration sequences (Table \ref{table_samples}).

\begin{table}
\caption{Sample sequence number collected in experiment.}\label{table_samples}
\begin{center}
\begin{tabular}{ccc}
\hline
Task         & Human Sequence \# & Robot Sequence \# \\ \hline
toy example  & 3                 & /                 \\
stack blocks & 2                 & 2                 \\
fold clothes & 2                 & 2                 \\ \hline
\end{tabular}
\end{center}
\end{table}
\begin{table}[]
\caption{Success rate of using state representation methods in two baseline controllers: UVS and REINFORCE. Spatial autoencoder has higher success rate since the mapping from task space to latent space is smooth and monotonous (shown in Fig. 5).}\label{control_success}
\begin{center}
\begin{tabular}{ccc}
\hline
Method              & UVS & REINFORCE \\ \hline
$\beta$-VAE         & 30\%                         & 40\%      \\
Spatial Autoencoder & 100\%                        & 100\%     \\ \hline
\end{tabular}
\end{center}
\end{table}

\subsection{Training}
Each task includes training on the above mentioned 4 state representation methods using the same data samples. For each training, different dimensions of the latent vector are also considered for the purpose of evaluating how compact should the state representation be.

\subsection{Evaluations on: State Dimension Reduction}
\subsubsection{How much compactness do we need?}
Selecting dimension of the latent space is a trade-off between compactness and keeping task information. The more compact the latent space is, the more efficient a controller can be designed~\cite{watter2015embed}. However, challenges may arise from over compact representations since some task relevant information is lost.

Given a sequence of demonstration images, we visualize each method's latent space on different dimension settings. We call this visualization the \textbf{task map} (Fig. 1). The x-axis defines the time line, while y-axis defines value of each element in the latent vector. A task map visualizes how latent vector changes along time line. Fig. 1 shows a representation with dimension size 100 is NOT much different from the one with size 50. But we can't conclude which one is better since one question is still unclear: what exactly is task relevant information that we should keep and what is non relevant that we can abandon?

\subsubsection{Disentangling time-varying and time-fixed factors}
Intuitively, during a task execution process, position changes (motions) are most relevant to a task, while other factors (e.g.,identity, shape, appearance) are not. Disentangling each of them is challenging~\cite{tulyakov2018mocogan}, how about simply divide them into two groups: \textbf{time-varying factors} and \textbf{time-fixed factors}. Because the majority of time-varying factors should be position changes, albeit appearance changes caused by motion is an exception. In experiments, we do find some elements in the latent space are changing along time, while some are fixed. This happens in methods e.g., VAE, $\beta$-VAE and spatial autoencoder. 

This phenomenon aligns with the three methods' theory background. VAE and $\beta$-VAE is basically mean-field variational Bayesian methods which assume independence of latent factors. They disentangle factors automatically, thus reveal some factors changing along the time while some are not. In Spatial Autoencoder, the last spatial softmax depicts positions represented in x and y coordinates. As a result, elements in this method's latent vector always come in pairs.

\begin{figure}
	\begin{center}
		\includegraphics[width=0.48\textwidth]{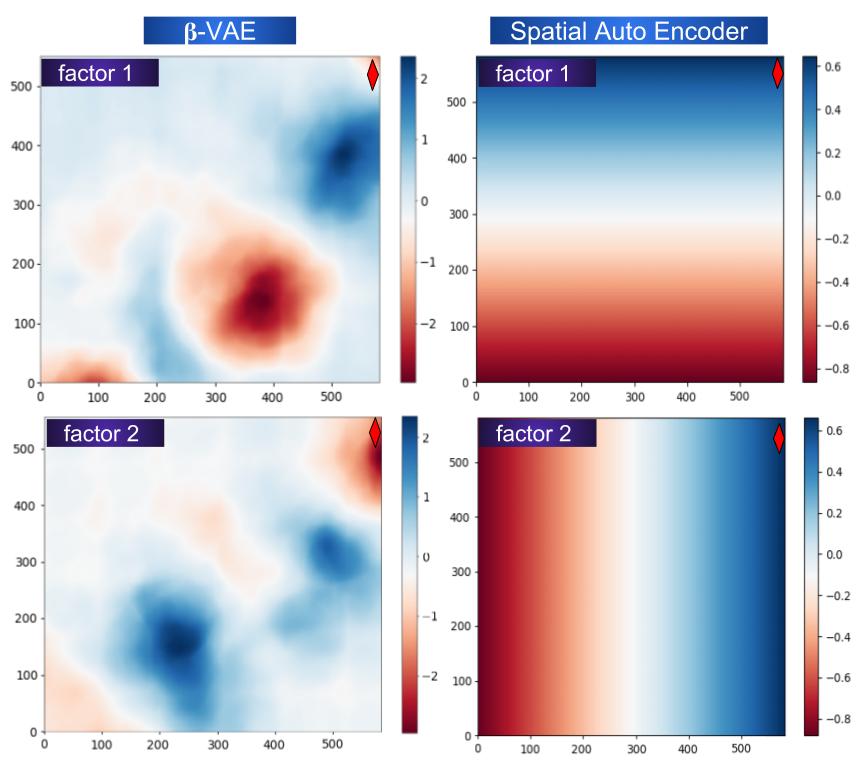} 
		\caption{The mapping from 2D task space to time-varying factors learned in state representation. We sample every point in task space of the toy example, and computes its corresponding $z^{v}_{t}$. Each point on this 2D plan corresponds to an image representing a state. The target state is marked using red diamond. For each mapping visualization, a total number of \textbf{580*580=336,400} images are generated and fed to different state representation method.}
		\label{fig:design_overview}
	\end{center}
\end{figure}
\subsubsection{Extracting time-varying factors}
As shown in Fig. 1, the time-varying factors in method $\beta$-VAE and Spatial autoencoder change smoothly along time line towards the target, while other time-fixed factors are almost vibrating around zero. It provides the possibility of only using time-varying factors in control. 

By extracting only time-varying factors in the latent vector, the state representation is more compact now. Fig. 3 shows how time-varying factors represent a certain trajectory pattern. Let's define an operator $\psi$ which extracts time-varying factors $z^{v}_{t}$ from a latent vector $s_{t}$:
\begin{equation}
\label{eq:time_varying}
z^{v}_{t} = \psi (s_{t}).
\end{equation} 

Fig. 1 also shows the number of time-varying factors increases as task DOF number increase. And it remains the same even if dimension of latent vector increases. For the toy example task, two time-varying factors are selected in $\beta$-VAE. In spatial autoencoder, four pairs of time-varying factors appear with the same value. We select the first pair in later experiments.

\subsubsection{Selecting hyper-parameter in $\beta$-VAE}
In experiments, we found different hyper-parameters ($\alpha$) in $\beta$-VAE greatly affects the result. According to its prior distribution assumption, variance of each factor in $z_{t}$ should be close to std variance. So we calculate a score based on how smooth the variance changes along time. The more smooth one returns a higher score. Fig. 4 shows an example of selecting the best parameter $\alpha$.

\begin{figure}
	\begin{center}
		\includegraphics[width=0.4\textwidth]{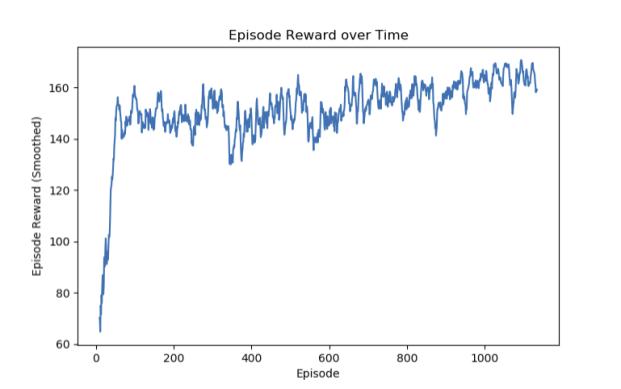} 
		\caption{Episode rewards over time in guided-REINFORCE training.}
		\label{fig:design_overview}
	\end{center}
\end{figure}

\subsection{Evaluation on: Controllability}
Given the time-varying factors $z^{v}_{*}$ of the target state, the control process of a policy $\pi(a_{t}|z^{v}_{t})$ is:
\begin{itemize}
    \item a) Get current image $I_{t}$, compute $z^{v}_{t}$ using eq. \ref{eq:11} and \ref{eq:time_varying}.
    \item b) Generate action $a_{t}$ from $\pi(a_{t}|z^{v}_{t})$. Execute action.
    \item c) If task not done, go to step a.
\end{itemize}

 Two baseline control methods are designed: Uncalibrated Visual Servoing (UVS[]) and REINFORCE.
 
\subsubsection{Baseline Control Method: UVS}
 UVS starts from an exploration manner to linearly map the observed $z^{v}_{t}$ changes to action changes. We follow the same process defined in~\cite{Ramirez2016} except for replacing error vector with our $z^{v}_{t}$. UVS controller is tested on all the three tasks. For the stack blocks task and fold clothes task, the state representation is re-trained using robot execution samples.
 
\subsubsection{Baseline Control Method: REINFORCE}
We use the same method as described in~\cite{jin2018robot} to estimate a reward function $r_{t}$ from demonstrated sequences. Image state in~\cite{jin2018robot} is substituted by $z^{v}_{t}$. In order to guarantee fast convergence, we designed \textit{Guided-REINFORCE} using gradient guidance towards the target. The objective here is to estimate an optimal policy $\pi^{*}(a_{t}|z^{v}_{t})$ which will result a maximum accumulated reward in finite time steps:
\begin{equation}\label{eq:pareto mle2}
  \begin{aligned}
\theta ^{*}=\argmax_{\theta}\mathbb{E}_{\pi_{\theta}}[\sum _{t} \gamma^{t}r_{t}]
 \end{aligned}
\end{equation}

$\pi(a_{t}|z^{v}_{t})$ is parameterized for continuous actions as:
\begin{equation}\label{eq:pareto mle2}
  \begin{aligned}
\pi_{\theta}(a_{t}|z^{v}_{t})=\mathcal{N}(\boldsymbol{\hat{a_{t}} + \mu _{t,\theta}}, \boldsymbol{\Sigma _{t,\theta}}),
 \end{aligned}
\end{equation}
where $\hat{a_{t}}$ is the guidance action which pull current state towards the target:
\begin{equation}\label{eq:pareto mle2}
  \begin{aligned}
\hat{a_{t}}=norm(\psi(z^{v}_{*}-z^{v}_{t})),
 \end{aligned}
\end{equation}
and the $norm$ operator normalize $\hat{a_{t}}$ to an action range limit. Process of updating policy gradient is the same as depicted in~\cite{sutton2018reinforcement}. The REINFORCE controller will plan small steps towards the target. As it involves real robot training which is impossible due to our time schedule, it's only tested in the toy example. Training curve on the toy example task can be found in Fig. 6.

\begin{figure}
	\begin{center}
		\includegraphics[width=0.4\textwidth]{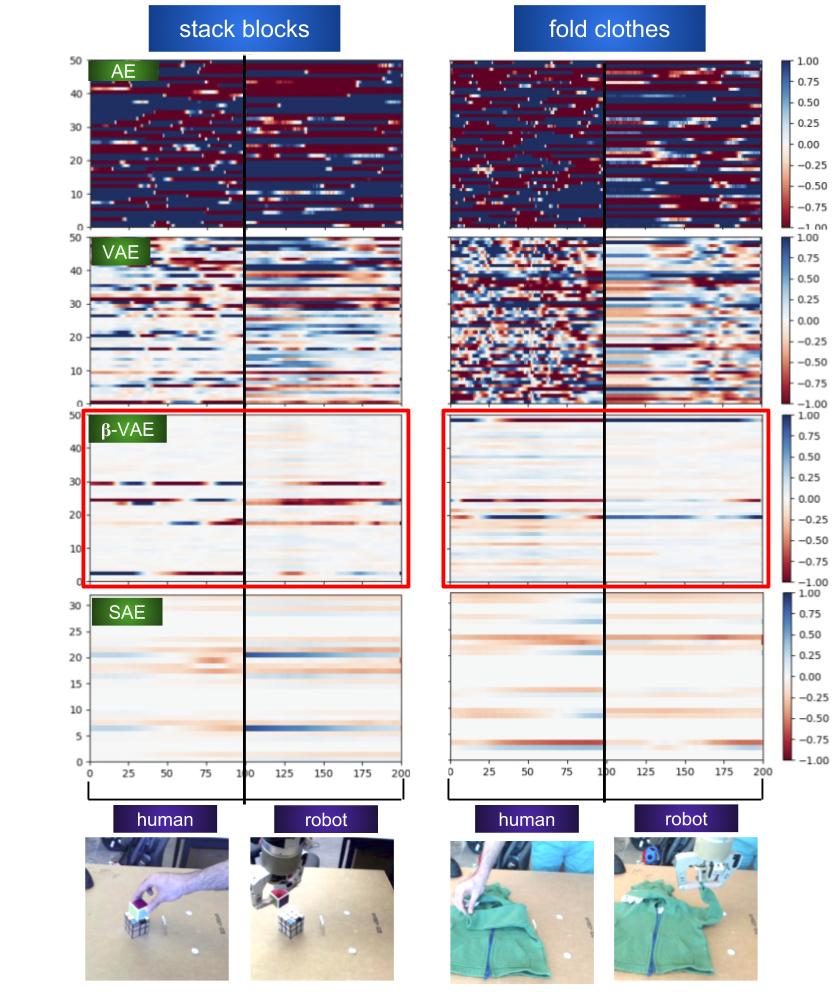} 
		\caption{Evaluation of transferring the learned state representation model to a robot case. The left map of each column shows the process that human performs the task. The right part of each column shows the process that robot performs the same task. For $\beta$-VAE, time-varying factor positions remain the same in all tasks. In stack blocks task, the last two time-varying factor reveals a same change pattern on both human and robot cases. In fold clothes task, this pattern can't be found. For spatial autoencoder, a coarse similarity pattern can be found but not clear enough.}
		\label{fig:design_overview}
	\end{center}
\end{figure}

\subsubsection{Results}
Directly control using time-varying factors learned from $\beta$-VAE and Spatial Autoencoder is challenging. Each task is executed 10 times on each state representation methods. Table \ref{control_success} shows the success rate on toy example task. Using UVS controller in stack blocks task and fold cloth task report no success since it's difficult to control with high DOF.

\begin{figure*}
	\centering
	\includegraphics[width=0.95\textwidth]{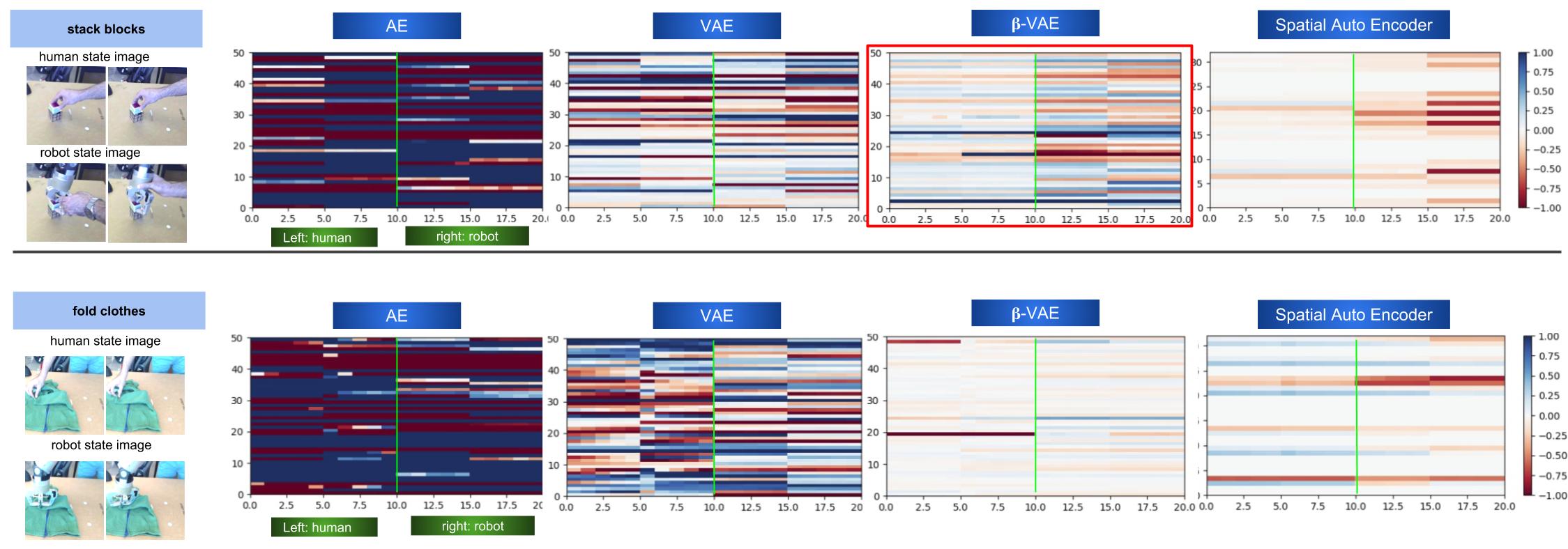}
	\caption{Evaluation of transferring the learned state representation model to a robot case, where human and robots both are in the same task state (e.g., the final state). For stack blocks task, two factors in $\beta$-VAE remain the same when human and robot holding the block at the final stage, however with different grapsing poses. Spatial autoencoder shows the same pattern but not clear enough. For fold clothes task, no such patterns are found in all the methods.}
	\label{fig:net_design}
\end{figure*}

\subsubsection{Why is it difficult?}
Time-varying factors smoothly change in the task space (Fig. 1). Furthermore, if different states in the task space result in different $z^{v}_{t}$, the whole configuration is global. Given a target state, it's promising to control towards the target.

However, the second property is not guaranteed. We sample every point in task space of the toy example, and computes its corresponding $z^{v}_{t}$. As shown in Fig. 5, both methods report smoothness. $\beta$-VAE captures more accuracy, however, it shows non-convex property in this mapping since different points in the 2D task space may correspond to the same $z^{v}_{t}$. Spatial autoencoder show both smoothness and monotonous properties which enable easier control.

In the $\beta$-VAE case, if the representation is augmented by adding a global configuration (i.e., xy coordinates of the object, joint states of the robot), it's promising to use global optimal control\cite{Peters2008} which will drive from any location towards the target.

\subsection{Evaluation on: Embodiment Problem}
In current researches, this ability is more investigated in how the learning process transferred (meta-learning~\cite{yu2018one}) in literature. We are more interested in the problem: what if we transfer the learned representation from human to robot? 
\subsubsection{Human and robot performing the same task}
Given a state representation model learned from human demonstration sequence, we evaluate over image sequences that a robot doing the same task. Fig. 7 shows the evaluation results.

\subsubsection{Visuallization of same task state}
Given a state representation model learned from human demonstration sequence, we evaluate human and robot performing the same task at the same task state (e.g, both are at the final task state). We wonder if the latent vectors are the same from human to robot. Results are shown in Fig. 8.

\section{Conclusions}\label{sec:conclusions}
In this paper, we evaluated different state representation methods in a robot hand-eye coordination task setting. Human demonstration provides an intuitive way to specify a robotic manipulation task. Imitation learning directly by watching human demonstrations~\cite{jin2018robot,yu2018one} is the closest to how human infants learn a task by observing adults and peers~\cite{rao2006cognition}. Understanding how states should be represented is important in understanding how to learn a task specification by watching demonstrations, as well as how to learn a control policy using image feedback. We hope this work could provide a little help in the above mentioned researches.

\addtolength{\textheight}{-2cm}   





\bibliographystyle{IEEEtran}
\bibliography{IEEEabrv,IEEEexample}

\end{document}